\documentclass[10pt,twocolumn,letterpaper]{article}

\usepackage{cvpr}              

\usepackage{graphicx}
\usepackage{amsmath}
\usepackage{amssymb}
\usepackage{booktabs}

\usepackage[pagebackref,breaklinks,colorlinks]{hyperref}

\usepackage[capitalize]{cleveref}
\crefname{section}{Sec.}{Secs.}
\Crefname{section}{Section}{Sections}
\Crefname{table}{Table}{Tables}
\crefname{table}{Tab.}{Tabs.}

\def\cvprPaperID{*****} 
\def\confName{CVPR}
\def\confYear{2022}

\begin{document}

\title{Addressing Bias Through Ensemble Learning and Regularized Fine-Tuning}

\author{Ahmed Radwan\\
KAUST\\
{\tt\small ahmedyradwan02@gmail.com}
\and
Layan Zaafarani\\
KAUST\\
{\tt\small layan.zaf@gmail.com}
\and
Jetana Abudawood\\
KAUST\\
{\tt\small jetana.abudawood@gmail.com}
\and
Faisal AlZahrani\\
KAUST\\
{\tt\small faiselalzahrani010@gmail.com}
\and
Fares Fourati\\
KAUST\\
{\tt\small fares.fourati@kaust.edu.sa}
}
\maketitle

\begin{abstract}
Addressing biases in AI models is crucial for ensuring fair and accurate predictions. However, obtaining large, unbiased datasets for training can be challenging. This paper proposes a comprehensive approach using multiple methods to remove bias in AI models, with only a small dataset and a potentially biased pretrained model. We train multiple models with the counter-bias of the pretrained model through data splitting, local training, and regularized fine-tuning, gaining potentially counter-biased models. Then, we employ ensemble learning for all models to reach unbiased predictions. To further accelerate the inference time of our ensemble model, we conclude our solution with knowledge distillation that results in a single unbiased neural network. We demonstrate the effectiveness of our approach through experiments on the CIFAR10 and HAM10000 datasets, showcasing promising results. This work contributes to the ongoing effort to create more unbiased and reliable AI models, even with limited data availability.

\end{abstract}

\section{Introduction}
\label{sec:intro}

Artificial Intelligence \cite{haenlein2019brief} is opening new solution doors every day, whether it be through computer vision \cite{forsyth2002computer}, natural language processing \cite{chowdhary2020natural} or more. Supervised learning is one of branches that has enabled us to go through these doors. It's seen as one of the most effective approaches as it uses the labeled data to let the model learn the underlying patterns and afterwards it tests the accuracy, it proved to be useful in many areas \cite{fourati2021artificial}. Such as agriculture \cite{kamilaris2018deep,fourati2021wheat}, medicine \cite{ching2018opportunities}, and software engineering \cite{flisar2019identification,javaid2016deep}.

However, despite its power, there are still challenges and limitations. Supervised learning's hunger for labeled data presents a significant challenge as obtaining the necessary labeled data is a costly and time-consuming process rather than obtaining unlabeled data. Another challenge supervised learning encounters is that it depends on having a large amount of data to achieve satisfactory \cite{ajiboye2015evaluating}. Therefore, using small data will make relying on the most straightforward methods, like training from scratch, unhelpful because it would not give us good results. To deal with this problem, we can use unsupervised learning \cite{hastie2009unsupervised}, semi-supervised learning \cite{zhu2005semi,van2020survey}, and fine-tuning \cite{yosinski2014transferable,swati2019brain}. 

An additional concern with Supervised Learning's hunger for data is bias. Bias happens when there is favouritism towards one entity more than the other, such as sexism, racism, etc. Bias is not limited to real-life situations – AI models face it too \cite{saunders2020reducing}. While we can't change people's biases in real life, our research focuses on fixing this problem in AI models. Trying to solve bias by changing data before training might sound simple, but it's not a good or easy way \cite{celis2020data}. Instead of going down that road, we will implement a different and generalized approach. We are using deep learning and transfer learning, tweaking the process with fine-tuning and regularization \cite{li2021improved}, all on a smaller, balanced dataset. This way seems more promising for dealing with bias effectively. 

Fine-tuning is one of the most common ways to implement transfer learning. Its key point is not to start the whole training process from scratch. Instead, it builds upon a pre-trained model \cite{chen2021pre} on a similar dataset. Fine-tuning tries to capture the essence of the training data based on the main characteristics of the large dataset from the pre-trained model, however, when the pre-trained model is biased, the fine-tuned model will capture the bias characteristics as well.

One other problem with fine-tuning is that it tends to over-fit \cite{ying2019overview} with a small amount of data. Dealing with overfitting on a small training dataset presents a significant challenge in transfer learning for domain adaptation. Therefore, we will implement a method that improves fine-tuning with regularization and to solve the bias issue as well.

Several papers have discussed different aspects of fine-tuning.
\cite{rozantsev2018beyond} The authors proposed a method with two architectures where one operates with the pre-trained data and the other with the personalized data where the weights are similar but not shared.
\cite{li2021improved} This paper addresses regularized fine-tuning for neural networks. The goal is to prevent the network from shifting from its pre-trained model and focus on the target task. This method is helpful because it balances the trade-off between regularization and generalization during fine-tuning.
\cite{saunders2020reducing} This paper addresses gender bias in Neural Machine Translation by leveraging transfer learning on a small gender-balanced dataset, achieving effective debiasing with reduced computational cost. A regularization term is added to the loss function when training a debiased model, aiming to maintain similarity between the debiased model and the original biased model's parameters. This term uses the Fisher information estimates that over samples from the biased data under the biased model to control the extent of regularization. This research paper uses a method to control bias in Natural Language Processing. We tend to reduce bias in a computer vision task by fine-tuning with regularization techniques.

Ensemble learning \cite{sagi2018ensemble} is a concept that combines techniques that merge multiple learning algorithms, often in supervised machine learning tasks. These algorithms, known as base-learners or inducers, take labeled data as input and generate models to generalize from this data. These models are then used to predict outcomes for new, unlabeled data points. Ensemble inducers can encompass various machine learning algorithms, like decision trees \cite{myles2004introduction}, neural networks \cite{lawrence1993introduction}, or linear regression \cite{gross2003linear} models. The central idea behind ensemble learning is that combining diverse models can mitigate the shortcomings of individual models, leading to improved overall predictive performance compared to a single model. In this paper, we will combine regularized fine-tuning with ensemble learning to remove bias.

Using only ensemble learning might result in slower training, but when combined with knowledge distillation, we have the opportunity to significantly improve the training process. knowledge distillation \cite{gou2021knowledge} is the process of transferring knowledge from a large model to a smaller one. In this method, a small "student" model is usually supervised by a large "teacher" model \cite{wang2021knowledge}. The core concept revolves around the idea that the smaller model imitates the larger model, aiming to achieve better outcomes.


\subsection{Contribution}
\begin{itemize}
    \item The paper presents a comprehensive approach to address bias in AI models using multiple techniques, even when working with a small dataset and a biased pre-trained model. This includes counter-bias training and ensemble learning.

    \item The proposed solution optimizes inference time by distilling the ensemble of models into a single unbiased neural network, enhancing model efficiency.

    \item The effectiveness of the approach is empirically demonstrated on real-world datasets, CIFAR10 and HAM10000, contributing to the ongoing effort to create more reliable and unbiased AI models, especially in situations with limited data availability.
    
\end{itemize}
\section{Problem Statement}
\label{sec:problem}

The primary goal of a neural network is to accomplish a specific objective, such as achieving high accuracy on seen and unseen data $(x,y)$. To achieve this goal, we define an  optimization problem aiming to minimize  a loss function $\mathcal{L}$ for the weights  $\theta$ of the model,
\begin{equation}
\min _{\theta}\mathbb{E}_{(x, y) \sim \mathcal{D}}[\mathcal{L}(\theta, x, y)] .
\end{equation}

Although this kind of training turns out to be successful in the standard regime, i.e., on sufficiently large data, it can still suffer from unfairness if the dataset is biased. We can avoid bias by ensuring our dataset is well-balanced.

Obtaining sufficient datasets for reliable model training might be time-consuming, and possibly expensive part \cite{bach2017learning} \cite{roh2019survey}, as it remains a significant challenge for AI engineers today. The main challenge is the potential for bias in the obtained dataset which ultimately leads to biased AI models. Bias is when there is one-sided favouritism towards one entity more than the other. With limited access to an extensive dataset, the potential effectiveness of our model's learning could be compromised, which leads to inaccurate predictions. One way to avoid this difficulty is to use a pre-trained model \cite{chen2021pre} that was trained on a large dataset and apply it to our data. However, using the pre-trained model directly can cause problems as the data it was trained on might differ from the datasets we have \cite{soekhoe2016impact} and could potentially be biased.

To overcome this problem, we can implement fine-tuning that presents an opportunity to reduce bias by allowing the model to start learning early while also adapting to our personalized dataset. However, the issue with fine-tuning is that small datasets could lead to overfitting \cite{ying2019overview}. Overfitting means the model performs better on the training data rather than the unseen testing data. This happens because the model will memorize the data instead of learning it\cite{fourati2021artificial}. Hence, we propose a method to improve fine-tuning using regularization \cite{barone2017regularization}. Additionally, we can improve the model's accuracy and fairness by including ensemble learning techniques to reduce bias.

\section{Proposed Solution}
\label{sec:solution}

\begin{figure*} [ht]
    \centering
    \includegraphics[width=\linewidth]{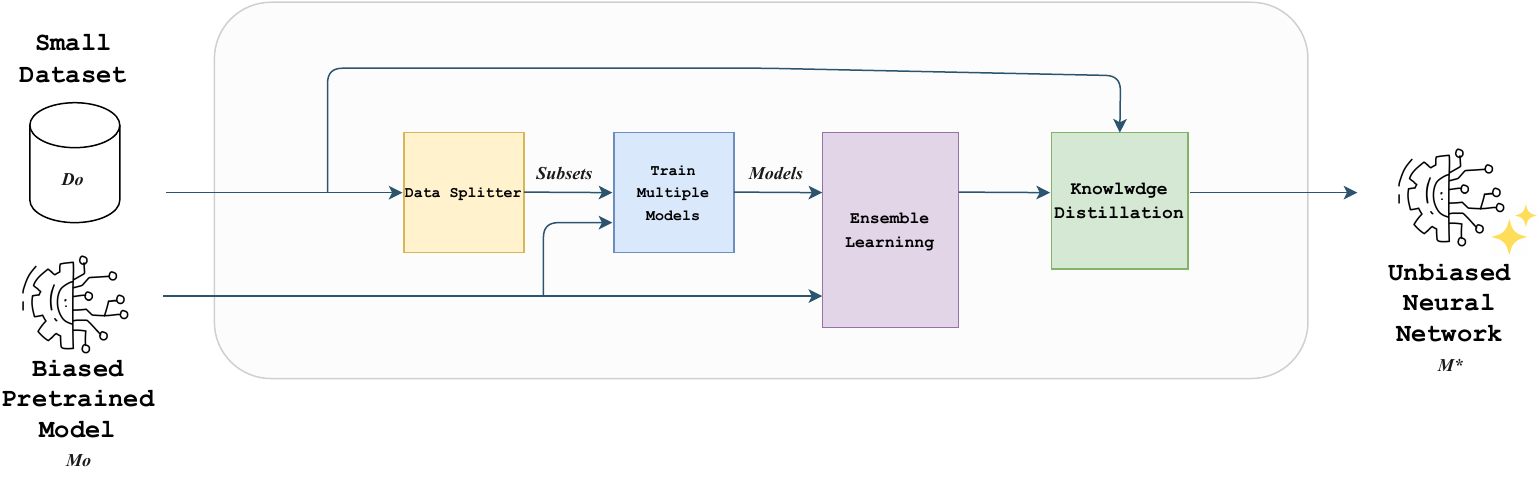} 
    \caption{Pipeline}
    \label{fig:plot}
\end{figure*}

Our solution requires two inputs: a small dataset and a pretraind model that is potentially biased. Given these inputs, we go through the pipeline shown in Fig1, The key point of the pipeline is ensemble learning, but to apply ensemble learning, we need a set of models. Therefore, we create $\mathcal{K}$ models derived from the pre-trained model using fine-tuning with regularization techniques. In the context of our study, we will view $\mathcal{K}$ as 2.


\begin{table*}[h]  
    \centering
    \caption{Pretrained Model with 10\% presence of missing labels}
    \label{tab:your_table_label}
    \begin{tabular}{lrrrrrrrrrrr}
\hline
 Model                   &   Plane &   Car &   Bird &   Cat &   Deer &   Dog &   Frog &   Horse &   Ship &   Truck &   Overall \\
\hline
 Initial Model           &   95.48 & 97.36 &  89.76 & 84.16 &  92.1  & 88.56 &  95.01 &   94.96 &  53.56 &   53.81 &      84.6  \\
 From Scratch            &    6.91 &  9.25 &   7.41 &  8.35 &  12.18 & 18.13 &  19.69 &    9.82 &  64.81 &   60.75 &      21.63 \\
 Reg. Fine-tuning &   80.15 & 92.34 &  81.13 & 63.08 &  70.6  & 79.22 &  68.38 &   79.47 &  75.16 &   62.55 &      75.39 \\
 Ensemble (Ours)               &   90.95 & 95.11 &  86.93 & 77.93 &  86.27 & 87.17 &  87.52 &   90.93 &  71.8  &   66.99 &      84.29 \\
\hline
\end{tabular}  
\end{table*}

\begin{table*}[h]  
    \centering
    \caption{Pretrained Model with 5\% presence of missing labels}
    \label{tab:your_table_label}
    \begin{tabular}{lrrrrrrrrrrr}
\hline
 Model                   &   Plane &   Car &   Bird &   Cat &   Deer &   Dog &   Frog &   Horse &   Ship &   Truck &   Overall \\
\hline
 Initial Model           &   96.58 & 98.84 &  90.6  & 86.41 &  92.63 & 83.51 &  93.96 &   94.97 &  20.58 &   21.3  &      78    \\
 From Scratch            &    7.26 & 10.19 &   5.64 & 16.23 &   6.28 & 10.54 &  10.91 &    6.67 &  52.16 &   50.07 &      17.57 \\
 Reg. Fine Tuning &   92.33 & 95.48 &  89.13 & 85.99 &  91.41 & 80.95 &  94.22 &   93.46 &  55.31 &   52.24 &      83.12 \\
 Ensemble (Ours)               &   86.16 & 90.71 &  86.85 & 85.3  &  90.59 & 79.19 &  93.96 &   92.2  &  67.23 &   61.47 &      83.44 \\
\hline
\end{tabular}  
\end{table*}

We split the data by creating multiple subsets for each model before training the $\mathcal{K}$ models. Each subset includes all entities from the missing classes in the initial pre-trained model and a small portion of entities from the rest of the classes. Hence, our strategy is to create biased subsets toward the poorly-represented classes in the original pre-trained model. Therefore, when the $\mathcal{K}$ models train on a biased dataset, they will also be biased.

To train those $\mathcal{K}$ models, we train some of them from scratch, and for some of them, we implement fine-tuning with regularization. We start from the pre-trained model's original parameters and then train it using our small dataset. Here is where the trick comes in: we modify how we measure the loss function to have a normal cross-entropy loss and add the difference between the current model we are training on and the trained one. To do this, we take the difference of the weights in each layer and then square the results. 

We also add another parameter $\lambda$ and define the following objective; 
\begin{equation}
\min _{\theta}\mathbb{E}_{(x, y) \sim \mathcal{D}}[\mathcal{L}(\theta, x, y) + \lambda \|\theta - \theta^*\|^2] .
\end{equation}
$\lambda$ regularizes the training and determines how far our model can drift from the initial model.

If $\lambda$ is large, the model will behave very similarly to the pre-trained model. But if $\lambda$ is small, the model will learn more about the personalized loss instead of the similarities of the two models. This rests under the assumption that we have a pre-trained model with a private dataset and a small dataset similar to the private dataset. Figuring out the perfect $\lambda$ parameter that determines how similar the model can be to the pre-trained model while maintaining a level of learnable weights from the dataset can be tricky. 

An addition to finding the similarities between the two models, we can add a second penalty regularization to reduce overfitting. We control its effect through beta such as:
\begin{equation}
\min _{\theta}\mathbb{E}_{(x, y) \sim \mathcal{D}}[\mathcal{L}(\theta, x, y) + \lambda \|\theta - \theta^*\|^2  + \beta\|\theta\|^2] .
\end{equation}
In our study, we will consider $\beta$ to be 0.

After training, we will have $\mathcal{K}$ models that are potentially biased. Therefore we can now apply ensemble learning to the models. For the sake of visioning the approach, let us imagine a dataset with 5 classes (1-5),  If the pre-trained model is biased toward some classes, e.g. (1-3), and the regularized models are now biased toward the other classes e.g. (4-5), ensemble learning will give the final, unbiased prediction correctly on all (1-5) classes. We do that by taking the average probabilities of each class from every model including the pre-trained model to conclude the unbiased prediction.

Our solution is finalized with knowledge distillation since multiple models could slow ensemble learning. The idea behind knowledge distillation is to combine the predictions from ensemble learning into the final single fair neural network.

\section{Experiments}
Our experiments encompassed several strategies to obtain optimal results. We conducted two experiments on CIFAR-10, where we created our own bias and tried to eliminate it with our proposed solution. The Experiments shown in Table 1 and Table 2 both showcase each class's accuracy on different models and the model's overall accuracy. The models are the initial pre-trained biased model, training from scratch, regularized fine-tuning, and our ensemble learning with models derived from the trained and fine-tuned with regularization. Table 1 shows where we initially trained the pre-trained model on a dataset that only had 10\% of entities from some classes (Ship and Truck). Table 2 is the same as Table 1 but with only 5\% of the entities from the missing classes (Ship and Truck). Now let us dive into the results:

\subsection{Regularized Fine Tuning}

\begin{figure} [ht]
    \centering
    \includegraphics[width=\linewidth]{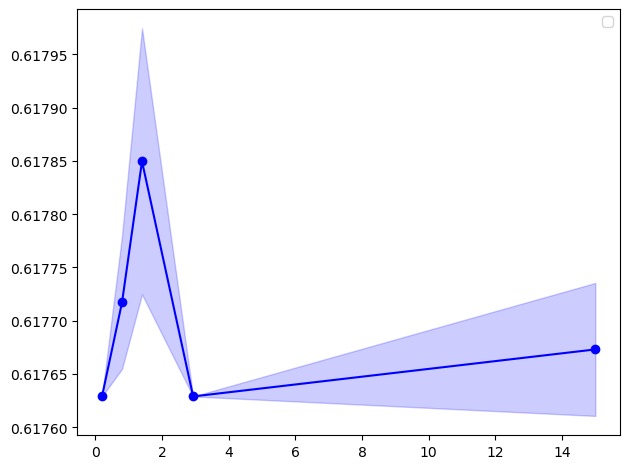} 
    \caption{Accuracy of Model for different Lambda values}
    \label{fig:plot}
\end{figure}

To fine-tune with regularization, we added a penalty term $\lambda$ to the loss function to counter the common problem of overfitting when dealing with small datasets. Before running the experiments, we searched for the best lambda value for effective regularization. Interestingly, we noticed that the accuracy for the missing classes improved as we increased this $\lambda$ value, as shown in Figure 2.

Table 1 displays these findings. It shows that using regularization during fine-tuning boosted the accuracy of those tricky classes. However, while the accuracy of these classes improved, the model's overall accuracy went down a bit to 75.39\%.

In Table 2, we tried a different approach. We trained the model using only 5\% of the problematic classes. Surprisingly, this change made those specific classes more accurate and lifted the overall model accuracy significantly. The overall accuracy jumped from 78\% to an impressive 83.12\%.

These experiments demonstrate how adding regularization techniques and smartly selecting classes for training can make the model better at handling the tough-to-predict classes and even improve the overall accuracy of the model.

\label{sec:experiments}

\subsection{Ensemble Learning}
To make our fine-tuning efforts more effective and considering the limitations of pre- and post-averaging, we tried out ensemble learning. This technique combines our regularized and original models, showcasing remarkable effectiveness.

As we can see in Table 1, when we used only 10\% of the images from the tricky classes (ship and truck), the accuracy for these classes shot up, leading to a significant increase in overall accuracy—from 75.39\% to a robust 84.29\%.

In Table 2, we went even leaner, using just 5\% of the images from those challenging classes (ship and truck). This time, the ship class accuracy increased from 55.31\% to 67.23\%, and the truck class accuracy went from 52.24\% to 61.47\%. Remarkably, we managed to keep the overall accuracy high, at 83.44\%.

The key to these impressive results was our decision to use logit summation instead of the usual voting method. This tweak played a significant role in achieving these excellent outcomes with our ensemble technique.

\subsection{Knowledge Distillation}
In pursuit of enhancing the performance of our complex model and addressing the challenges posed by a limited dataset, we turned our attention to the technique of knowledge distillation. This method, rooted in model compression and transfer learning, proved to be a pivotal strategy in achieving improved outcomes while also addressing concerns related to computational efficiency. However, our exploration took an unexpected turn. Building on the spectacular successes of our ensemble, we incorporated the sum of logits into our loss function for training the compressed model. Contrary to our expectations, this endeavour failed, resulting in a significant drop in accuracy to 35.77

\section{Conclusion}
In this paper, our proposed approach combines fine-tuning with regularization, ensemble learning, and knowledge distillation to address biases in AI models when faced with small datasets. By leveraging the strengths of these techniques, we aim to reduce bias and improve the accuracy of predictions. Our experiments on both the CIFAR10 and HAM10000 datasets have shown promising results, highlighting the potential of our approach to enhance model fairness and performance. Further research and optimization in this direction could lead to even more significant advancements in reducing bias and improving AI model predictions.



{\small
\bibliographystyle{ieee_fullname}
\bibliography{egbib}
}

\appendix

\section{Experimental Details}
\subsection{CIFAR10}
In our cifar10 \cite{krizhevsky2010convolutional} experiment, we used a dataset with ten classes containing 50,000 training images alongside 10,000 test images where we sampled 25 images from each of the ten classes to have a total of 250 images for the personalized dataset. We aim for one “pre-trained model” and multiple “Personalized models”. The pre-trained model's dataset consists of all classes, with classes 8 and 9 having only 5\% and in another experiment 10\% number of images compared to classes 1-7 which comes to 40,456 images for training (5,000 images for classes 1-7 and 250 images for classes 8, 9, and 500 images on the second experiment). The balanced test dataset containing 10,000 images was used for testing. On the other hand, our personalized model's dataset consists of all ten classes with 4,000 images; we removed images of class 8, and 9 completely. And split the dataset in two halves (same classes but different images), and added the images of classes 8, 9 back to the two subsets creating two final subsets with 2,395 images each, thus creating a personalized dataset with a small bias towards class 8, 9 that is relatively small compared to the original dataset.
We have tried implementing many different model architectures but with experimenting we found that using Densenet121 \cite{huang2017densely} gave the most accurate and better results. As for the hyperparameters, all models were trained for 20 epochs using a learning rate (lr) of 0.0001. A learning rate scheduler was used to adjust the lr every five steps reducing it by 0.9 each time. We also used momentum for purposes of optimization with a rate of 0.5. As an optimizer, the Stochastic Gradient Descent (SGD) \cite{bottou2012stochastic} was chosen, and we made sure that model hyperparameters were the same for all models in our setup. This setup of the hyperparameters helped the model learn better and improve its performance.
We used a wide range of methods to carefully identify the best strategy for getting great results. Our thorough testing process included a variety of tactics. To demonstrate, we first averaged a set of global models, and then trained the resulting model on a balanced dataset. This iterative approach used a range of lambda values, starting at 0, to cover the whole range of conventional fine-tuning. Additionally, we pursued the method of regularized fine-tuning on individual models, which later conducted averaging on them. Lastly, instead of taking the average, we tried ensemble learning to benefit the most from the diversity in the models.
For the best outcomes, our implementation tried to enhance model generality. Due to our small dataset, traditional fine-tuning was unsuccessful, which raised the risk of overfitting. We fixed this by adding a lambda penalty to the loss function. This reduced overfitting by balancing model complexity and data quality. The learning process was led by the lambda penalty, producing a more flexible and reliable solution.

Recognizing that neither pre- nor post-averaging produced satisfactory outcomes, we embarked on an exploration of ensemble learning, which yielded the most promising results. Our ensemble methodology, comprising both regularized and global models, showcased remarkable performance.

Improving our approach has been a key concern in the framework of our experimental undertakings. We have carefully improved our pipeline structure through several thorough trials to improve the overall performance of our models. This project consists of a number of crucial parts, each of which has been carefully planned to increase the effectiveness of our strategy. Our first step is to choose a pre-trained model to serve as the basis for our research. Next, we carefully enrich our dataset to include examples from the previously underserved groups to address the shortcomings of the pre-trained model. Our dataset is then divided into two separate subsets, each serving a particular purpose: that includes complete photos of the classes that were previously missing and half images from the last classes. To guarantee fair representation and variety, this partitioning approach is carried out in a randomized manner. Both subsets are used to train two different models in the training phase, which benefits from the use of regularized fine-tuning approaches to improve the models' adaptability and generalization abilities. Our strategy achieves its maximum potential when ensemble learning is combined, where the information of the pre-trained model, the first model, and the second model comes together to produce a final prediction that is more precise and durable. This combination of methodologies, which includes ensemble learning, unbiased subset division, targeted dataset enhancements, and pre-trained models, promises to overcome the drawbacks of individual models and ultimately result in significant improvements in predictive accuracy.

\subsection{Skin Cancer}
The HAM10000 dataset \cite{tschandl2018ham10000} was used as another application to implement our method. The dataset has seven classes, and each class resembles a skin lesion. 10000 images were used for training, and 1500 images were used for testing, where we sampled all images from all seven classes. The pre-trained model's dataset consists of 5 classes with 6405 images, where we used 5124 for training. The remaining 1281 images were used for testing. On the other hand, the personalized model's dataset consists of all seven classes with a total of 805 images; we used only 402 for training and the remaining 402 for testing. 
Several model architectures were also tested on this dataset, but yet again, Densenet121 \cite{huang2017densely}  gave us the best results. As for the hyperparameters of this application, the initial model was trained for ten epochs using a learning rate (lr) of 0.001. A learning rate scheduler was used to adjust the lr every five steps reducing it by 0.9 each time. We also used momentum for purposes of optimization with a rate of 0.5. And as an optimizer, the Stochastic Gradient Descent (SGD) \cite{bottou2012stochastic} was chosen. Furthermore, the personalized model was trained for 50 epochs using a learning rate (lr) 0.001. A learning rate scheduler was used to adjust the lr every five steps reducing it by 0.9 each time. We also used momentum for purposes of optimization with a rate of 0.5. And as an optimizer, the Stochastic Gradient Descent (SGD) was chosen.

\end{document}